\documentclass[runningheads]{llncs}
\usepackage{graphicx}
\usepackage{amsmath}
\usepackage{amssymb}
\usepackage{booktabs}
\usepackage{subcaption}
\usepackage{multirow}
\usepackage{placeins}
\usepackage{pifont}
\newcommand{\xmark}{\text{\ding{55}}}
\usepackage{makecell}

\usepackage[pagebackref,breaklinks,colorlinks]{hyperref}
\usepackage{algorithm, algorithmic}

\usepackage{tikz}
\usepackage{comment}
\usepackage{amsmath,amssymb} 
\usepackage{color}

\usepackage[accsupp]{axessibility}  

\begin{document}
\pagestyle{headings}
\mainmatter
\def\ECCVSubNumber{100}  

\title{Self-Supervised Representation Learning for Adversarial Attack Detection} 


%
\author{Yi Li \and
Plamen Angelov \and
Neeraj Suri}
%
\authorrunning{Y. Li et al.}
%
\institute{Computing and Communications, Lancaster University, UK\\
\email{y.li154@lancaster.ac.uk}}
\maketitle

\begin{abstract}
Supervised learning-based adversarial attack detection methods rely on a large number of labeled data and suffer significant performance degradation when applying the trained model to new domains. In this paper, we propose a self-supervised representation learning framework for the adversarial attack detection task to address this drawback. Firstly, we map the pixels of augmented input images into an embedding space. Then, we employ the prototype-wise contrastive estimation loss to cluster prototypes as latent variables. Additionally, drawing inspiration from the concept of memory banks, we introduce a discrimination bank to distinguish and learn representations for each individual instance that shares the same or a similar prototype, establishing a connection between instances and their associated prototypes. We propose a parallel axial-attention (PAA)-based encoder to facilitate the training process by parallel training over height- and width-axis of attention maps. Experimental results show that, compared to various benchmark self-supervised vision learning models and supervised adversarial attack detection methods, the proposed model achieves state-of-the-art performance on the adversarial attack detection task across a wide range of images.
\end{abstract}

\section{Introduction}
Given an image potentially perturbed by an attack algorithm, the goal of adversarial attack detection is to distinguish between adversarial and normal samples using the differences between them. Adversarial attack detection is an important security topic applicable in real-world applications such as autonomous driving systems, object detection, medical image processing, and robotics \cite{seg}\cite{NLP}\cite{context}\cite{multihead} among many others. With recent advancements in deep learning, several neural-network-based approaches have been proposed for adversarial attack detection \cite{detecting}\cite{IJCNN}. These networks and approaches are predominantly trained in a supervised manner, where a large number of labeled adversarial and normal samples are provided as input to neural networks. The model is then trained to reconstruct the corresponding clean sample and compare it with the input sample to provide the detection result. Consequently, supervised learning-based adversarial attack detection approaches suffer from three main drawbacks.

Firstly, human-imperceptible adversarial attacks on images are challenging to label manually. This process can be time-consuming and may introduce errors, particularly when the annotator lacks familiarity with the task. Secondly, the trained adversarial attack detection models may need to be deployed in previously unseen conditions, including novel attack algorithms and datasets. Consequently, there is a strong likelihood of a mismatch between the training and testing conditions. In such cases, we lack the ability to leverage recorded test data to improve the model's performance in the unseen test setting. Thirdly, prototype-based adversarial attack detection methods \cite{UNICAD}\cite{cvprw} estimate an object's category (e.g., cats or dogs) as the prototype. These methods calculate the degree of similarity between new data samples and autonomously chosen prototypes to classify images as adversarial or normal samples. However, each prototype may potentially consists of multiple instance samples, which often leads to a neglect of the rich intrinsic semantic relationships between prototypes of individual objects in images. For example, while the model may be trained on some tank images, it may struggle to classify new tanks or entirely new classes of objects when faced with previously unseen types of tanks.

To overcome these drawbacks, our contributions are summarized as follows:

$\bullet $ We propose a self-supervised representation learning framework aimed at extracting feature representations for the downstream task, i.e., adversarial attack detection. Building upon pixel mapping \ref{3.2} and contrastive estimation in \ref{3.3}, in \ref{3.4}, we propose a discrimination bank to distinguish individual instances for each prototype from the embedding space. We demonstrate that the instance-wise feature maps capture richer information compared to the prototype-based approach, resulting in performance improvements.

$\bullet $ In \ref{3.5}, we propose a parallel axial-attention (PAA)-based encoder to split the 2-D attention map into two 1-D sub-attention maps, one for height and one for width. Unlike the original axial-attention approach \cite{axial}, PAA can be simultaneously trained on two GPU devices, enabling parallel calculations of the attention maps and facilitating the training process.

$\bullet $ We demonstrate the effectiveness of our proposed methods by comparing them to state-of-the-art pre-trained models and existing adversarial attack detection methods across a diverse range of images.

\section{Related Works}
\subsection{Self-Supervised Learning}
Self-supervised learning aims to develop effective feature representations without the need for large annotated datasets, thereby addressing the annotation bottleneck, which is one of the primary challenges in the practical deployment of deep learning today. Recent studies have shown a growing interest in self-supervised learning, particularly after Yann LeCun's keynote address at the AAAI 2020 conference \cite{lecun}. Generally, self-supervised learning can be categorized into three main approaches: predictive \cite{pred}, generative \cite{IJCNN2024}, and contrastive learning \cite{contr}. For instance, He et al. proposed masked autoencoders (MAE) to mask random patches of input images and reconstruct the missing pixels \cite{mae}. 


\subsection{Contrastive Learning}
Contrastive learning aims to develop low-dimensional representations of data by contrasting similar and dissimilar samples \cite{contr}. It encourages the learning of feature representations with both inter-class separability and intra-class compactness, which can be highly beneficial for network learning.

One type of contrastive loss function used for self-supervised learning, noise-contrastive estimation (InfoNCE), employs logistic regression to distinguish the target data from noise \cite{nce}. Ding et al. use prototypes derived from contextual information to better explore the intrinsic semantics of relations \cite{pcl1}. However, the application of contrastive learning in pixel-level change detection remains a challenging and relatively unexplored area.
\subsection{Axial-attention} 
Wang et al. introduce axial-attention in their work \cite{axial}, which splits 2D self-attention into two 1D self-attentions. This approach reduces computational complexity and enables attention operations within larger or even global regions. As another variety, Li et al. generate time and frequency sub-attention maps by calculating attention maps along the time and frequency axes of speech spectra \cite{ushaped}. To more efficiently highlight local foreground information, Li et al. propose group parallel axial-attention (GPA) to solve medical image segmentation task \cite{GPA}. It's worth noting that these studies are primarily conducted in supervised settings, which necessitate labeled data during model training.

 \section{Self-Supervised Representation Learning}
Our proposed solution is built on a progression of novel contributions that combine to build the detection framework. As depicted in Fig 1, the innovations consist of a proposal for pixel mapping (\ref{3.2}) that facilitates mapping input images into the embedding space through a pair of data transformations. Section \ref{3.3} develops a contrastive estimation approach for adversarial attack detection. As the core idea of our contributions, Section \ref{3.4} provides the discrimination bank, which preserves each instance corresponding to its associated prototype. Finally, we provide an overview of each network component, with a particular focus on our proposed PAA in \ref{3.5}.

\begin{figure*}[ht]
  \centering
   \includegraphics[width=12.2cm, height=2.2cm]{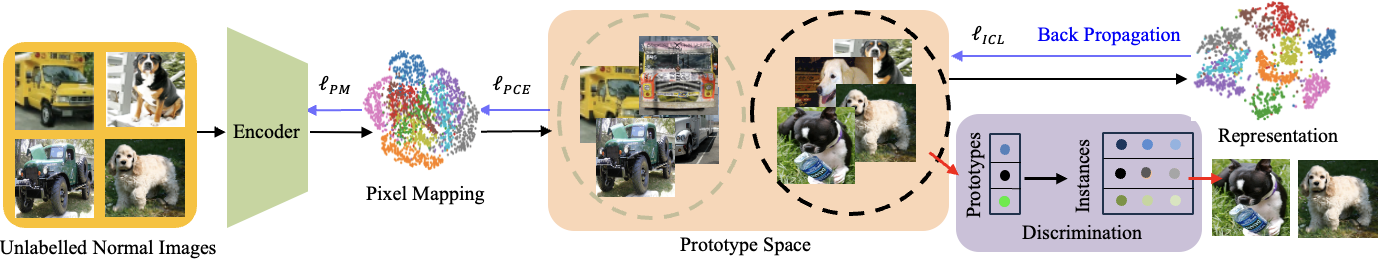}
   \caption{Self-supervised representation learning framework}
   \label{fig:onecol}
\end{figure*}
\subsection{Preliminaries}
Given a training set $X=\{x_1,x_2,...,x_n\}$ of $n$ image samples, we aim to learn the instance-wise feature representation $Z=\{z_1,z_2,...,z_n\}$ of unlabelled normal images for the downstream task, i.e., adversarial attack detection. As a contrastive learning method, InfoNCE \cite{infonce} achieves this objective by optimizing a contrastive loss function, defined as:
\begin{equation}
\mathcal{L}_{\text {InfoNCE }}=\sum_{i=1}^n-\log \frac{\exp \left(v_i \cdot v_i^{\prime} / \tau\right)}{\sum_{j=0}^r \exp \left(v_i \cdot v_j^{\prime} / \tau\right)}
\end{equation}
where $V=\{v_1,v_2,...,v_n\}$ is the embedding vectors for $n$ instances, and $v_{i}^{\prime}$ is the $i$-th positive embedding. Moreover, $v_{j}^{\prime}$ includes one positive embedding and $r$ negative embeddings for other instances, and $\tau$ is a dynamic hyper-parameter. In this work, we estimate the prototypes by replacing $v_i^{\prime}$ in equation (1) with prototype $p_i^{+}$, enabling the proposed discrimination bank establish the connection between the prototype space and each instance.

\subsection{Pixel Mapping} \label{3.2}
As the first major component of the encoder, a PAA-based network with parameter $\theta$ is exploited to transform $X$ to feature vectors $V=\{v_1,v_2,...,v_I\}$, such that $V$ best describes $X$. Different from previous work (e.g., InfoNCE loss), we propose a novel pixel mapping loss with data augmentation, $\mathcal{L}_{\text {PM}}$, to learn an invariant representation of $x_i$ by minimizing the risk $\sum_i \mathcal{L}\left(x_i, v_i ; \theta\right)$. To achieve that, we use a pair of transformations, denoted as $t$ and $s$, in some set of transformations $\mathcal{T}$ (e.g. geometric transformations, color transformations, etc.) to $x_i$, to produce the augmentation as $x_i^{t_i}$ and $x_i^{s_i}$. We define this process as $V= f_{PM}(X)$ with the loss as: 
\begin{equation}
\mathcal{L}_{\text {PM}}=-\log \frac{\exp \left(f_{PM}\left(x_i^{t_i}\right)^T \cdot f_{PM}\left(x_i^{s_i}\right) / \tau\right)}{\sum_{b=1}^B \exp \left(f_{PM}\left(x_b^{t_b}\right)^T \cdot f_{PM}\left(x_i^{s_i}\right) / \tau\right)}
\end{equation}
where $T$ and $B$ are the transpose symbol and batch size, respectively. It is highlighted that all the embeddings in the loss function are L2-normalized \cite{l2}. While previous data augmentation studies \cite{t} have shown that the choice of transformation techniques plays an important part in self-supervised representation learning, most previous works do not give much consideration to the individual choice of $t_i$ and $s_i$ on pairs of images, which are simply uniformly sampled over $\mathcal{T}$. Therefore, in the proposed pixel mapping technique, we aim to overcome this limitation and select the optimal transformation algorithm for each sample $x_i$. To achieve this, we select transformation algorithms that maximize the risk defined by the loss $\mathcal{L}^{\mathrm{PM}}$:
\begin{equation}
\left\{t_i, s_i\right\}=\underset{\left\{t_i, s_i\right\} \in \mathcal{T}}{\arg \max } \sum_{i=1}^{n} \mathcal{L}_{\text {PM}}\left(x_i^{t_i}, x_i^{s_i} ; \theta, \mathcal{T}\right)
\end{equation}
In the proposed pixel mapping technique, we prioritize the difference between $t_i$ and $s_i$ for each image over their absolute values.
\subsection{Prototype-wise Contrastive Estimation} \label{3.3}
We assume that the observed data $x_i$ are related to latent variable $P=\{p_i\}$ which denotes the prototypes of the data. We aim to find a network parameter that maximizes the log-likelihood function of the observed $n$ samples by a prototype-wise contrastive estimation (PCE). To achieve that, we use the local peaks of the density \cite{prot} as the prototype, in other words, the most representative data samples of $X$. The loss, namely $\mathcal{L}_{\text {PCE }}$, is defined as:
\begin{equation}
\mathcal{L}_{\text {PCE }}=\frac{1}{\left|\mathcal{M}\right|} \sum_{p_i^{+} \in \mathcal{M}}-\log \frac{\exp \left(v_i \cdot p_i^{+} / \gamma\right)}{\sum_{p_i^{-} \in \mathcal{N}} \exp \left(v_i \cdot p_i^{-} / \gamma\right)}
\end{equation}
where $\mathcal{M}_i$ and $\mathcal{N}_i$ are prototype collections of the positive and negative samples, respectively. As aforementioned, inspired from previous supervised learning work \cite{proto}\cite{simdnn}, we find different levels of concentration distributes around each prototype embeddings. Therefore, we exploit $\gamma$ as the concentration level around the prototype $p^{m}$ within the $m$-th cluster as:
\begin{equation}
\gamma=\frac{\sum_{i=1}^n\left\|p^{m}-v_i^{m}\right\|_2}{n \log (n+\beta)}
\end{equation}
where the momentum features are denoted as $\{v_i^{m}\}_{i=1}^{n}$ within the same cluster as a prototype $p$. We set a smooth parameter $\beta$ to ensure that small clusters do not have an overly-large $\gamma$. In the proposed prototype clustering, $\gamma$ acts as a scaling factor on the similarity between an embedding $v$ and its prototype $p$. With the proposed $\gamma$,the similarity in a loose cluster (larger $\gamma$) are down-scaled, pulling embeddings closer to the prototype. On the contrary, embeddings in a tight cluster (smaller $\gamma$) have an up-scaled similarity, thus less encouraged to approach the prototype. Therefore, learning with PCE yields more balanced clusters with similar concentration.
\subsection{Instance-Wise Contrastive Learning} \label{3.4}
The core of our method lies in establishing a connection between prototype and instance features to facilitate instance clustering. This approach enables accurate classification of unseen and attacked samples based on the learned representation. To accomplish this, drawing inspiration from the recent success of memory banks \cite{mb}, we introduce a discrimination bank for clustering instances that share a common prototype. Initially, we create $K$ independent discrimination banks to enhance instance discrimination across clusters. Similar to a memory bank, the discrimination bank aids in contrastive learning, leveraging extensive data to acquire robust representations. We assume a contrastive set $J_i$ for the $t$-th bank $A_{t}$ as:
\begin{equation}
J_i= \{z_{i}^{\prime} \mid z_{i}^{\prime} \in A_t \forall t \in[1, C] \}
\end{equation}
where $z_{i}^{\prime}$ is the estimated representation of $x_{i}$. Specifically, for each training batch with $B$ samples and $M$ prototypes, our discrimination memory is built with size $M\times B\times D$, where $D$ is the dimension of pixel embeddings. The ($p^{m}$, $b$)-th element in the discrimination memory is a $D$-dimensional feature vector obtained by average pooling all the embeddings of pixels labeled as $p^{m}$ prototype in the $b$-th batch. To update the discrimination bank, we enqueue each instance to the nearest prototype and add the new one in each back propogation cycle:
\begin{equation}
\mathcal{L}_{\text {ICL}}=\frac{\exp (\cos(v_i,z_{i}) \cdot \cos \left(v_i , p_{i}^{m} / \phi\right))}{\sum_{z^{\prime} \in A_t}\sum_{j=0}^r \exp (\cos(v_i,z_{j}^{\prime}) \cdot \cos \left(v_i , p_{j}^{m} / \phi\right)) \cdot J_i}
\end{equation}
where cos$(\cdot,\cdot)$ is the cosine similarity between a pair of representations. The concentration level of $\mathcal{L}_{\text {ICL}}$ is presented as $\phi$ and estimated similar as $\gamma$ in (4) but we replace $v_{c}^{\prime}$ to $z_{c}^{\prime}$. With the loss, we discriminate representations belongs to the same bank. To discover the underlying concepts with unique visual characteristics, we infer their decision boundaries by reducing the visual redundancy among clusters, namely maximising the visual similarity of samples within the same clusters and minimising that between clusters. Concretely, as the representation of samples with different pseudo labels are stored independently in the discrimination bank, they can be taken as anchors to describe their corresponding clusters. The overall cost-function used to train the MAE is now a combination of the above loss terms with hyper-parameters $\lambda_1$ and $\lambda_2$:

\begin{equation}
\mathcal{L}=\mathcal{L}_{\text {PM}}+\lambda_1\cdot\mathcal{L}_{\text {PCE}}+\lambda_2\cdot\mathcal{L}_{\text {ICL}}
\end{equation}


\subsection{Parallel Axial-attention-Based Encoder} \label{3.5}
Our encoder has four major components for each learning objective:

1. For $\mathcal{L}_{\text {PM }}$ in Eq. (2), we set the temperature $\tau$ as 0.1. To transform $X$ into embeddings $V$, we use a ResNet-50 \cite{rc} as the backbone. However, different from conventional ResNet families, we transform it to a parallel axial-attention (PAA)-ResNet by replacing the convolutional layer in the residual bottleneck block by two parallel multi-head axial-attention layers (one for height-axis and the other for width-axis). Additionally, two Conv1D layers are included to shuffle the features, as illustrated in Fig. 2.

\begin{figure}[ht]
  \centering
   \includegraphics[width=12cm, height=3.1cm]{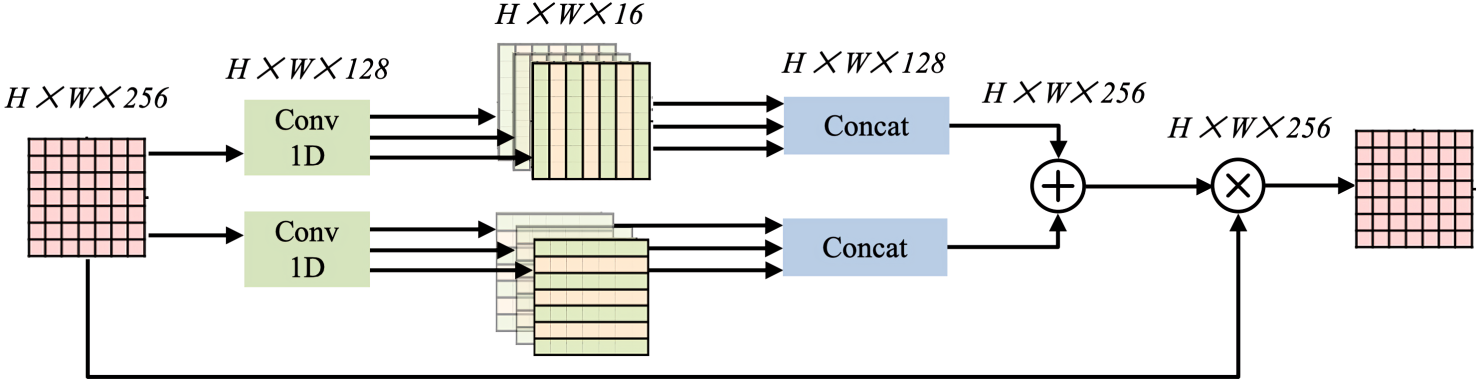}
   \caption{Parallel axial-attention (PAA) block.}
   \label{fig:axial}
\end{figure}

2. Prototype-wise contrastive estimation (PCE) is achieved by two Conv1D layers with ReLU. PCE is only applied during training and is removed at inference time. Thus it does not introduce any changes to the segmentation network or extra computational cost in deployment.

3. Discrimination bank consists of two parts that store prototype and instance embeddings, respectively. For each prototype, we set the maximum size of the instance queue as 10. The discrimination bank is discarded after training.

4. Instance-wise contrastive learning is achieved by two Conv1D layers with ReLU and faiss \cite{faiss} for efficient instance clustering. Similar as PCE, ICE is removed in the test stage.
\section{Experiments}
\subsection{Datasets and Attacks}
We randomly select 50,000 images from ImageNet \cite{imagenet} and 10,000 images from ImageNet Large Scale Visual Recognition Challenge (ILSVRC) \cite{ILSVRC} for the training and validation, respectively. As aforementioned, we evaluate the competitor and proposed models with unseen datasets. In the test stage, we extensively perform experiments on several public datasets, including ImageNet-R \cite{imager}, Canadian Institute For Advanced Research-10 (CIFAR-10) \cite{cifar10}, and Microsoft Common Objects in Context (COCO) \cite{coco}. In each above-mentioned dataset, we randomly select 10,000 images to evaluate the detection performance of models.

We select seven attack algorithms \cite{pgd}\cite{fgsm}\cite{deepfool}\cite{bim}\cite{cw}\cite{jsma}\cite{SSAH} in the test stage because they are robust to novel adversarial attack detection and defense techniques. Parameters of all the seven attacks are shown in Table 1.

\begin{table}[htbp!]
\centering
\caption{Parameters of seven adversarial attacks}
\begin{tabular}{cc}
\hline
Attack & Parameters\\
 \hline
FGSM & $\epsilon$=0.008 \\
PGD & $\epsilon$=0.01, $\alpha$=0.02, Steps=40 \\
SSAH & $\alpha$=0.01 \\
DeepFool & Steps=20 \\
BIM & $\epsilon$=0.03, $\alpha$=0.01, Steps=10 \\
CW & C, Kappa=2, Steps=500, learning rate=0.01 \\
JSMA  & $\gamma$=0.02 \\
 \hline 
\end{tabular}
\end{table}
\subsection{Backbones and Competitors}
As aforementioned we use ResNet-50 as the encoder's backbone without pre-training. Moreover, various backbones (i.e., HRNet \cite{hrnet} and Xception \cite{Xception}) in supplementary experiments to validate the proposed algorithm and compare to ResNet-50.

In addition, the proposed method is evaluated and compared to state-of-the-art competitor models. Firstly, we reproduce four supervised adversarial attack detection techniques \cite{simdnn}\cite{sensor}\cite{toler}\cite{detecting} as the original implementations in the literature but with same data as the proposed method. Secondly, we use five pre-trained self-supervised models \cite{tico}\cite{mae}\cite{mugs}\cite{unic}\cite{dino} which are state-of-the-art in image processing tasks. Finally, we reproduce two state-of-the-art unsupervised adversarial attack detection methods \cite{rebut}\cite{rebut2}.

\subsection{Implementation Details}
In the experiment, we build the PAA-ResNet-S from ResNet-50 \cite{rc} whose last fully-connected layer outputs a 128-D and L2-normalized feature. As aforementioned, to achieve that, we replace the convolutional layer in the residual bottleneck block by two parallel multi-head axial-attention layers (one for height-axis and the other for width-axis). We multiply all the channels by 1.5 and 2, resulting in PAA-ResNet-{M, L}, respectively. We always use 8 heads in multi-head attention blocks \cite{atyn}. In order to avoid careful initialization of weights ($W_Q$, $W_K$, $W_V$) and location vectors ($r^{q}$, $r^{k}$, $r^{v}$), we use batch normalizations \cite{bm} in all attention layers. To evaluate and compare the adversarial attack detection accuracy, we use the detection rate (DR).

The proposed model is trained by using the SGD optimizer with a weight decay of 0.0001, a momentum of 0.9, and a batch size of 256. We train the networks for 200 epochs, where we warm-up the network in the first 20 epochs by only using the pixel-mapping loss. The initial learning rate is 0.03, and is multiplied by 0.1 at 120 and 160 epochs. In terms of the hyper-parameters, we set $\tau=0.1$, $\beta=10$, $r=16000$, $\lambda_1=1$ and $\lambda_2=1$. All the experiments are run on the High End Computing (HEC) Cluster with Tesla V100 GPUs.


\section{Results}
\subsection{ImageNet}
The learned representation is evaluated on adversarial attack detection task over the ImageNet-R dataset. Table 2 shows the results, each of them is the average of 70,000 experiments (10,000 images$\times$7 attacks). 

\begin{table}[htbp!]
\centering
\caption{Comparison on the ImageNet-R dataset.}
\begin{tabular}{c|cc|cc|cc}
\hline
 & \multicolumn{2}{c|}{Configuration}  &\multicolumn{2}{c|}{Computational Cost}  &\multicolumn{2}{c}{DR ($\%$)}\\ 
\hline
Method & Supervised & Pre-training  &Para.& Latency (ms)  &Clean& Attacked\\ 
 \hline
TiCo \cite{tico} &\xmark &\checkmark  &178.0 M & 160.3&79.5 & 65.9 \\
MAE \cite{mae} &\xmark & \checkmark  & 307.8 M & 104.5&88.4 & 72.0\\
Mugs \cite{mugs} &\xmark &\checkmark & 303.5 M&185.4 & 89.2 & 72.1 \\
Unicom \cite{unic} & \xmark &\checkmark  & 303.5 M&268.9 & 91.4 & 80.5 \\
DINOV2 \cite{dino} & \xmark & \xmark & 1.0 B &572.0 & 92.5 & 82.8 \\
\hline
ESMAF \cite{sensor} &\checkmark & \xmark  & 171.0 M & 196.2&69.8 & 53.7\\
TS \cite{detecting} &\checkmark & \xmark  & 36.5 M & 45.5&78.0 & 57.6\\
sim-DNN \cite{simdnn} &\checkmark & \checkmark  & 663.9 M & 227.1&79.2 & 60.3\\
DTBA \cite{toler} &\checkmark & \checkmark & 554.2 M & 208.4&84.2 & 62.3\\
\hline
TLC \cite{rebut} & \xmark &\checkmark & 27.5 M & 6.9  & 83.5 & 71.7\\
SimCat \cite{rebut2} & \xmark &\checkmark & 27.8 M & 7.2 &85.1 & 72.9\\
\hline
 \textit{PAA-ResNet-S} & \xmark & \xmark & {\bfseries 26.8 M}&{\bfseries 6.2} & 92.0 & 83.1 \\
  \textit{PAA-ResNet-M} & \xmark & \xmark &  35.4 M& 8.7  &  93.5 & 85.2\\
  \textit{PAA-ResNet-L} & \xmark &\xmark  & 45.9 M&  10.1  & {\bfseries 93.8} & {\bfseries 86.8}\\
 \hline 
\end{tabular}
\end{table}
Table 2 shows the averaged adversarial attack detection performance of the proposed method as compared to \cite{simdnn}\cite{sensor}\cite{toler}\cite{detecting}\cite{tico}\cite{mae}\cite{mugs}\cite{unic}\cite{dino}\cite{rebut}\cite{rebut2} using the ImageNet-R dataset. From Table 2, it can be observed that: (1) In all the evaluated models, the proposed PAA-ResNet-L achieves 93.8$\%$ and 86.8$\%$ for clean and attacked images detection, respectively, which offers the best effectiveness. (2) Compared to state-of-the-art models, the proposed PAA-ResNet family requires the least computational cost and inference latency in the training stage because the model utilizes a parallel computation, which makes it feasible in real-world applications.
\subsection{COCO $\&$ CIFAR-10}
We assess the learned representation over CIFAR-10 and COCO. Tables 3 $\&$ 4 show the results, each of them is the average of 70,000 experiments (10,000 images$\times$7 attacks).


\begin{table}[htbp]
    \begin{minipage}{.5\linewidth}
        \centering
        \caption{Comparison on CIFAR-10. }
\begin{tabular}{cccc}
\hline
Models & Clean ($\%$) & Attacked ($\%$)\\
 \hline
 TiCo \cite{tico} & 81.4 & 78.0\\
MAE \cite{mae} & 89.9 & 74.2 \\
Mugs \cite{mugs} & 90.5 & 73.7 \\
Unicom \cite{unic} & 92.6 & 84.1 \\
DINOV2 \cite{dino} & 94.3 & 86.7  \\
\hline
ESMAF \cite{sensor} & 73.8 & 56.4 \\
TS \cite{detecting} & 89.7 & 59.5 \\
sim-DNN \cite{simdnn} & 82.0 & 65.7 \\
DTBA \cite{toler} & 87.0 & 74.1 \\
\hline
TLC \cite{rebut}  & 84.9 & 72.4\\
SimCat \cite{rebut2}  &88.0 & 77.3\\
 \hline 
 
\textit{PAA-ResNet-S} & 92.7 & 84.4 \\
\textit{PAA-ResNet-M} & 94.1 & 87.8 \\
\textit{PAA-ResNet-L} &{\bfseries 94.8} &{\bfseries 89.0} \\
\hline
\end{tabular}
        
    \end{minipage}%
    \begin{minipage}{.5\linewidth}
        \centering
        \caption{Comparison on COCO.}
\begin{tabular}{cccc}
\hline
Models & Clean ($\%$) & Attacked ($\%$)\\
 \hline
 TiCo \cite{tico} & 78.9 & 67.3\\
MAE \cite{mae} & 88.9 & 73.5 \\
Mugs \cite{mugs} & 89.0 & 73.3 \\
Unicom \cite{unic} & 90.2 & 82.8 \\
DINOV2 \cite{dino} &{\bfseries 91.7} & 83.9  \\
\hline
ESMAF \cite{sensor} & 75.4 & 55.6 \\
TS \cite{detecting} & 76.7 & 56.8 \\
sim-DNN \cite{simdnn} & 80.6 & 62.2 \\
DTBA \cite{toler} & 85.3 & 68.8 \\
\hline
TLC \cite{rebut}   & 80.8 & 71.5\\
SimCat \cite{rebut2} &82.6 & 70.1\\
 \hline 
\textit{PAA-ResNet-S} & 90.9 & 83.7 \\
\textit{PAA-ResNet-M} & 91.5 & 84.9 \\
\textit{PAA-ResNet-L} &{\bfseries 91.7} &{\bfseries 85.6} \\
\hline 
\end{tabular}
        
    \end{minipage}
\end{table}

On both datasets, our models show strong detection performance: accuracy improves considerably with the proposed algorithm. Additionally, our results outperforms both the self-supervised and supervised results by large margins on clean images detection.
\subsection{Diagnostic Experiment}
In this section, we conduct the ablation study to evaluate the effectiveness of each contribution.

\subsubsection{Contrastive Learning}
In this section, we compare the effectiveness of each proposed contrastive learning loss to previous contrastive learning techniques and conduct the ablation study on the ImageNet-R dataset. Moreover, we compare our contrastive losses to similar losses in \cite{nce}\cite{pcl1}. Each result is the average of 70,000 experiments (10,000 images$\times$7 attacks).
\vspace{-1em}
\begin{table}[htbp!]
\caption{Ablation study of the three contributions in the proposed method.}
\centering
\begin{tabular}{lcc}
\hline
$\hspace{0.8cm}$Ablation Settings &Clean ($\%$) &Attacked ($\%$) \\
\hline
PAA-ResNet-L  & 60.1 & 52.3 \\
 \hline
$+\mathcal{L}_{\text {NCE}}$ \cite{nce}  & 72.0  & 63.5 \\
$+$Data Augmentation \cite{t} & 65.6 & 50.1 \\
$+\mathcal{L}_{\text {PM}}$ & 76.7 & 68.9 \\
 \hline 
 $+\mathcal{L}_{\text {S2Z}}$ \cite{pcl1} & 79.7 & 68.6 \\
 $+\mathcal{L}_{\text {PCE}}$ & 82.5 & 72.4 \\
 $+\mathcal{L}_{\text {PM}}+\mathcal{L}_{\text {PCE}}$ & 90.4 & 82.8 \\
 \hline 
 $+\mathcal{L}_{\text {PCE}}+\mathcal{L}_{\text {ICL}}$ & 89.5 & 82.0 \\
 $+\mathcal{L}_{\text {PM}}+\mathcal{L}_{\text {PCE}}+\mathcal{L}_{\text {ICL}}$ &{\bfseries 93.8} &{\bfseries 86.8} \\
 \hline 
\end{tabular}
\end{table}
\vspace{-1em}

We first perform experiments to validate the design of our pixel mapping with a PAA-ResNet-L as the baseline. As shown in Table 3, additionally considering data augmentation leads to a substantial performance gain (i.e., 5.4 $\%$), compared with $\mathcal{L}_{\text {NCE}}$.

We next investigate the effectiveness of $\mathcal{L}_{\text {PCE}}$. On the one hand, $\mathcal{L}_{\text {PCE}}$ boosts the performance based on $\mathcal{L}_{\text {PM}}$ (i.e., 68.9$\%\rightarrow$82.8$\%$). On the other hand, when we replace $\mathcal{L}_{\text {PCE}}$ to a prototype-based loss $\mathcal{L}_{\text {S2Z}}$ \cite{pcl1}, the proposed $\mathcal{L}_{\text {PM}}+\mathcal{L}_{\text {PCE}}$ achieves a higher score (i.e., 82.8$\%$) is achieved, which confirms the efficiency of the proposed loss terms. 

We then presents a comprehensive examination of $\mathcal{L}_{\text {ICL}}$. Compared with $\mathcal{L}_{\text {PM}}+\mathcal{L}_{\text {PCE}}$, $\mathcal{L}_{\text {ICL}}$ brings a substantial performance gain (i.e., 4.4 $\%$). Moreover, the performance slightly drops (i.e., 82.8$\%\rightarrow$82.0$\%$) when $\mathcal{L}_{\text {ICL}}$ replace $\mathcal{L}_{\text {PM}}$ within combining $\mathcal{L}_{\text {PCE}}$.

\subsubsection{Parallel Axial-attention}
We conducted experiments to demonstrate the trade-off between performance improvement and network depth, specifically, varying the number of proposed PAA blocks. It's important to note that each PAA block consists of two parallel sub-blocks, i.e., height and width attentions. Furthermore, we compared the performance of different backbones, such as ResNet-50, HRNetV2-W48 \cite{hrnet}, and Xception-71 \cite{Xception}. Additionally, we assessed the performance improvements resulting from the inclusion of axial-attention \cite{atyn} and PAA. Fig. 3 (a) and (b) present these results, with each data point being an average of 70,000 experiments (10,000 images of ImageNet-R $\times$ 7 attacks).
\begin{figure}[htbp!]
\centering
\includegraphics[width=12cm, height=2.7cm]{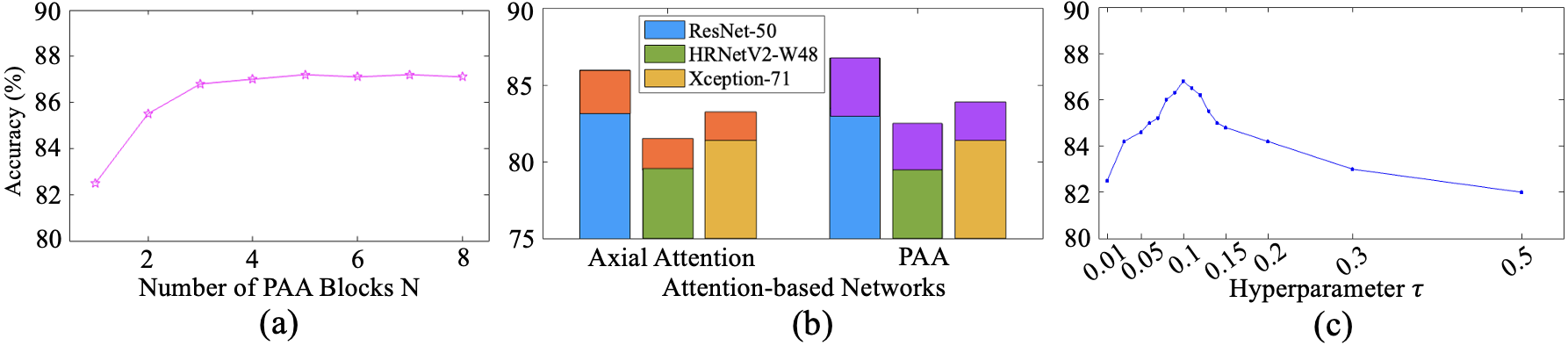}
\caption{Ablation study for (a) number of parallel axial-attention (PAA) blocks and (b) attention blocks to different backbones (c) hyperparameter $\tau$. The red and purple rectangles represent accuracy improvements when axial attention and PAA are added to the backbones, respectively.}\centering
\end{figure}

Fig. 3(a) compares the number of PAA blocks against detection accuracy on ImageNet-R. The results indicate that N=3 offers the best trade-off, validating the chosen implementation setting. Fig. 3(b) presents the experiment results of networks with the axial-attention and PAA. It can be observed that: (1) Among the three backbones without added attention blocks, ResNet-50 outperforms HRNetV2-W48 and Xception-71. (2) Both axial-attention and PAA enhance detection performance based on three backbones. Particularly, the proposed PAA provides a greater performance boost compared to the axial-attention block, with an improvement from 2.1$\%$ to 2.9$\%$ on average across the three backbones. 

\subsubsection{Hyper-parameter}
An ablation study of hyper-parameters is conducted using the PAA-ResNet-L on the ImageNet-R dataset. Fig. 3(c) shows downstream attack detection accuracy when $\tau$ varies from 0 to 2. Besides, $\beta$ impact is evaluated in Fig. 4(a). More ablation studies are performed to evaluate the loss term hyper-parameters $\lambda_1$ and $\lambda_2$ in Fig. 4(b). Each data point being an average of 70,000 experiments (10,000 images $\times$ 7 attacks).

\begin{figure}[htbp!]
\centering
\includegraphics[width=12cm, height=2.9cm]{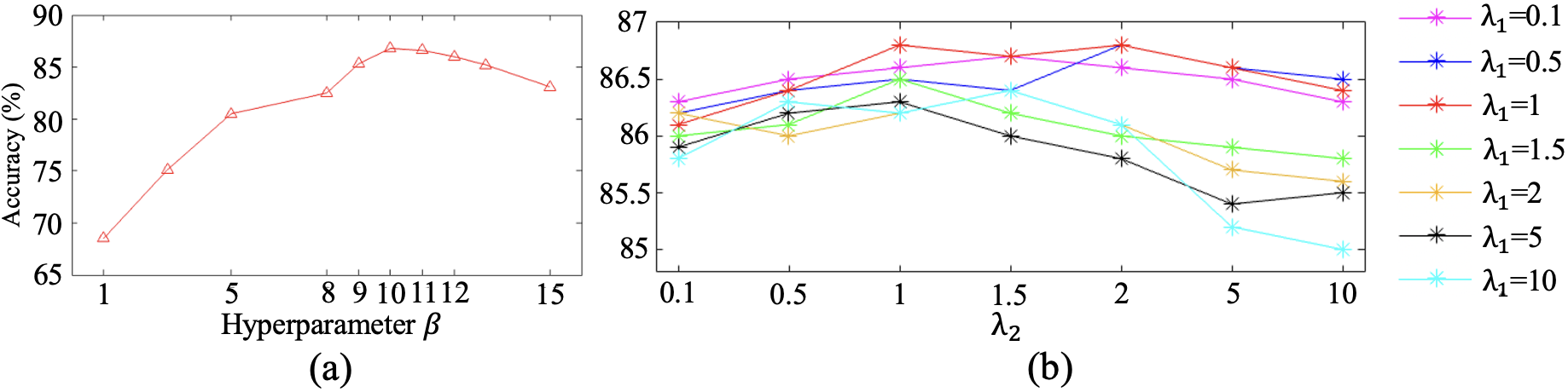}
\caption{Ablation study for (a) hyper-parameter $\beta$ and (b) $\lambda_1 \&\lambda_2$.}\centering
\end{figure}

As Fig. 3(c) shows, detection accuracy starts to increase with $\tau$ = 0.01 and reaches its peak around $\tau$ = 0.1, but performance is fairly stable for 0.03 $\le\tau \le$ 0.2. Additionally, Fig. 4(a) suggests the hyper-parameter $\beta$ = 10. Fig. 4(b) presents detection accuracy against $\lambda_1$ and $\lambda_2$. There is no significant accuracy drop even when the importance of loss terms is significantly weighted, such as by as much as tenfold that of $\mathcal{L}_{\text {PM}}$ ($\lambda_1,\lambda_2$ = 10). This demonstrates that the features derived from each loss terms contribute positively to the learning process.



\subsection{Robustness Evaluation}
We perform experiments to evaluate the robustness of the supervised and self-supervised models on adversarial attack detection. For fair comparison, we fine-tune pre-trained self-supervised models \cite{unic}\cite{mae}\cite{dino}\cite{mugs} using the same training data as the supervised models on the corresponding dataset. Table 6 shows the detection accuracy results (in $\%$), each of them is the average of 70,000 experiments (10,000 images $\times$ 7 attacks). Apart from aforementioned datasets, we introduce one more dataset, i.e., Canadian Institute For Advanced Research-100 (CIFAR-100) \cite{cifar10}.

\begin{table}[]
\centering
\caption{Adversarial attack detection performance (clean / attacked images) on seen and unseen datasets.}
\scalebox{0.90}{
\begin{tabular}{c|cc|cc|cc}
\hline
Training     & \multicolumn{2}{c|}{ImageNet-R} & \multicolumn{2}{c|}{ILSVRC} & \multicolumn{2}{c}{CIFAR-100} \\
\hline
Test     & ImageNet-R      & CIFAR-10     & ILSVRC     & CIFAR-100  & CIFAR-100 & ImageNet-R \\
\hline
MAE \cite{mugs} & 89.4 / 74.1 & 89.0 / 73.4 & 91.0 / 79.2 & 89.1 / 73.0 & 90.4 / 74.3 & 85.4 / 70.5 \\
Mugs \cite{mugs} & 89.8 / 74.0 & 89.1 / 73.6 & 91.8 / 78.1 & 89.4 / 74.5 & 90.9 / 75.0 & 86.2 / 71.1 \\
Unicom  \cite{unic} & 91.9 / 82.7& 91.0 / 80.4 & 94.7 / 88.5 & 92.0 / 81.1 & 93.3 / 82.7 & 89.3 / 77.9 \\
DINOV2   \cite{dino} & 93.4 / 84.5 & 92.4 / 81.7 & 96.2 / 90.0& 93.4 / 82.6 & 95.1 / 84.0 & 90.5 / 79.4 \\
\hline
ESMAF \cite{sensor} & 88.1 / 72.3 & 75.5 / 59.7 & 90.1 / 78.7 & 77.8 / 62.6 & 89.5 / 72.6 & 74.2 / 59.8 \\
TS \cite{detecting} & 90.2 / 75.8 & 79.6 / 66.2 & 92.6 / 79.8 & 82.3 / 68.1 & 91.2 / 79.8 & 83.7 / 66.6 \\
sim-DNN  \cite{simdnn} & 90.1 / 77.4 & 81.5 / 70.6 & 93.4 / 82.5 & 81.9 / 71.2 & 92.8 / 82.4 & 83.6 / 65.9 \\
DTBA    \cite{toler} & 92.2 / 85.2 & 85.3 / 76.9 & 96.0 / 90.3 & 86.8 / 78.2 & 94.7 / 83.1 & 88.2 / 69.9 \\
\hline
\textit{PAA-ResNet-L} & {\bfseries 93.5 / 87.9} & {\bfseries 92.9 / 85.7} & {\bfseries 97.1 / 90.5} & {\bfseries 94.2 / 87.0} & {\bfseries 96.0 / 87.6} & {\bfseries 92.1 / 83.4} \\             
\hline
\end{tabular}}
\end{table}
It can be observed that the adversarial attack detection accuracy on unseen datasets is lower compared to the seen dataset utilized in both the training and test stages, primarily due to differences in distributions of datasets. However, when compared to supervised learning-based methods \cite{sensor}\cite{detecting}\cite{toler}\cite{simdnn}, the proposed SSL representation learning method experiences relatively less performance degradation.

\subsection{Visualization}
As qualitative analysis, Fig. 5 presents the t-distributed stochastic neighbour embedding (t-SNE) visualisation of PAA-ResNet-L trained with different losses. Compared to the representation learned by $\mathcal{L}_{\text {PM}}$, the representation learned by $\mathcal{L}_{\text {ICL}}$ forms more separated clusters, which also suggests representation of lower entropy. In Fig. 5(b), it can be observed that the feature embeddings within a single prototype are not separable. However, when the discrimination bank is added in Fig. 5(c), individual instances become separated. This demonstrates that the proposed methods can learn discriminative feature representations that generalize well for adversarial attack detection across various attack algorithms.

\begin{figure}[htbp]
    \centering
    \begin{subfigure}[b]{3.5cm}
        \centering
        \includegraphics[width=3.5cm]{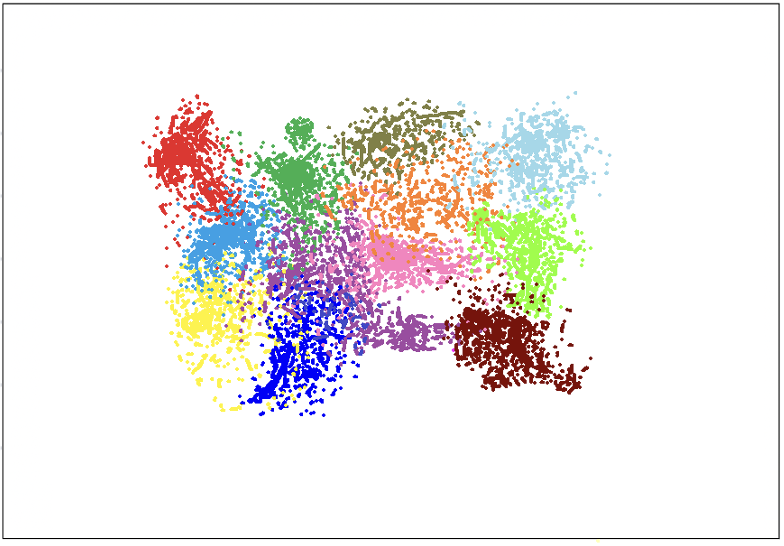} 
        \caption{Pixel mapping}
    \end{subfigure}
    \hfill
    \begin{subfigure}[b]{3.5cm}
        \centering
        \includegraphics[width=3.5cm]{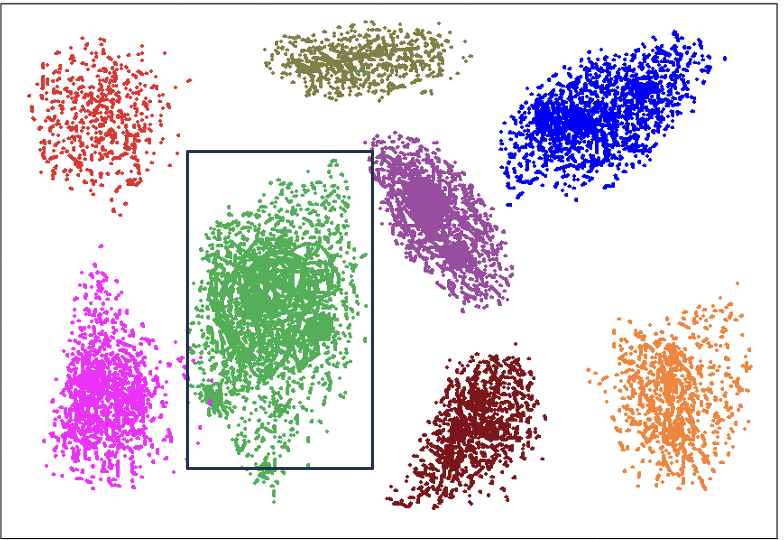} 
        \caption{PCE}
    \end{subfigure}
    \hfill
    \begin{subfigure}[b]{3.5cm}
        \centering
        \includegraphics[width=3.5cm]{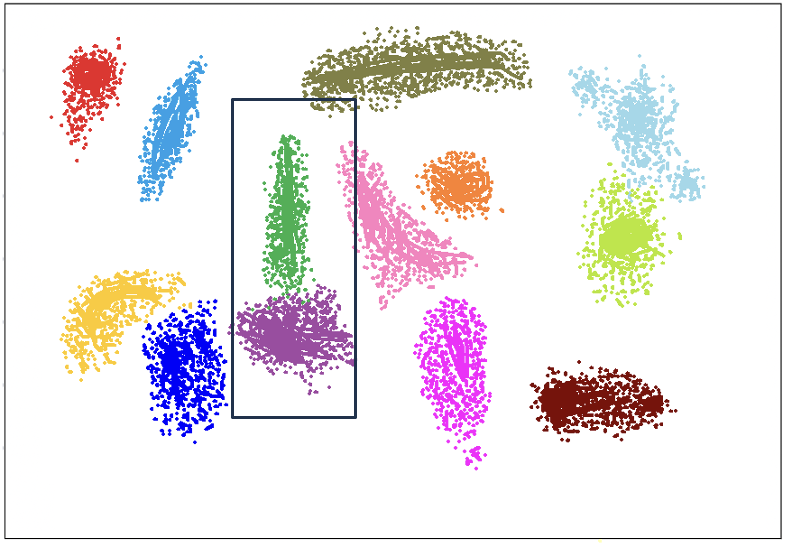} 
        \caption{ICL}
    \end{subfigure}
    \caption{t-SNE feature visualization of the model with (a) $\mathcal{L}_{\text {PM}}$; (b) $\mathcal{L}_{\text {PM}}+\mathcal{L}_{\text {PCE}}$; (c) $\mathcal{L}_{\text {PM}}+\mathcal{L}_{\text {PCE}}+\mathcal{L}_{\text {ICL}}$.}
\end{figure}
\section{Discussion and Conclusion}
We enumerate these advantages of this work below:

1. In the training stage, we only require access to unlabelled normal image samples. Therefore, unlike supervised adversarial attack detection methods, we do not need paired training data, i.e., adversarial and normal images. Moreover, unlike supervised learning models, the proposed model extracts the underlying sub-class structure from the augmented data distribution and encodes it into the embeddings. This guarantees the robustness of the downstream task.

2. The proposed loss $\mathcal{L}_{\text {PM}}$ considers the difference between two transformation techniques for each image $x_i$ more than their absolute values. These transformed images create more challenging positive pairs and $\mathcal{L}_{\text {PM}}$ produces harder negative pairs by accounting for all images in a batch during the optimization, enabling the learning of richer features. 

Adversarial attacks result in a deviation of the constructed graphs of adversarial examples from the prototypes of their correct classes, creating challenges for accurate classification based on representations. Despite this, the distributions between the reconstructed image and the prototypes of target classes remain significantly different. The proposed $\mathcal{L}_{\text {PCL}}$ correctly encourages representations to be closer to their assigned prototypes because the encoder would learn the shared information among prototypes, and ignore the individual noise that exists in each prototype. The shared information is more likely to capture higher-level feature knowledge.

By using the proposed discrimination bank, the encoder preserves the connection between the instance and the associated prototype. This enables the learning of richer features from samples of a single class. Therefore, the proposed method outperforms other contrastive learning methods in the literature. 

3. PAA blocks capture richer feature information, enabling models to seamlessly integrate local and global feature representations, thereby enhancing their ability to provide local-global feature information for downstream tasks. Additionally, PAA blocks simplify computational complexity by allowing height and width attentions to be simultaneously trained on parallel GPU devices to facilitate the training. Furthermore, PAA blocks can be easily implemented on various backbones and consistently demonstrate promising results.



While the concept of parallel axial-attention structure has been introduced in previous work \cite{GPA}, it's important to note that the algorithm in our work is fundamentally different from the literature. In \cite{GPA}, the feature maps are reshaped to $(H\times C)\times W$ and $(W\times C)\times H$ in two parallel attentions blocks, respectively. However, in this work, we only focus on one dimension, either width or height, in each sub-PAA block. Therefore, the computational cost is further reduced.

In this paper, we have proposed a self-supervised representation learning approach for adversarial attack detection, offering an effective alternative to traditional supervised pipelines. We establish a connection between prototype and instance features through the use of a discrimination bank, thereby enriching the information available to enhance the proposed model's ability to detect adversarial attacks. Our evaluation with different datasets and attacks has demonstrated the robust performance of the proposed method on unseen datasets. Additionally, PAA-ResNet models offer faster inference speeds compared to competitive pre-trained models, making them feasible for potential real-world applications.
\section*{Acknowledgment}
This work is supported by the UKRI Trustworthy Autonomous Systems Node in Security/EPSRC Grant EP/V026763/1.

This work is also supported by ELSA – European Lighthouse on Secure and Safe AI funded by the European Union under grant agreement No. 101070617. Views and opinions expressed are however those of the author(s) only and do not necessarily reflect those of the European Union or European Commission. Neither the European Union nor the European Commission can be held responsible.

\clearpage
%
%
\bibliographystyle{splncs04}
\bibliography{paper}
\end{document}